\documentclass[format=acmsmall, review=false, screen=true]{acmart}

\usepackage{booktabs} 

\usepackage[ruled,linesnumbered]{algorithm2e} 

\SetAlFnt{\small}
\SetAlCapFnt{\small}
\SetAlCapNameFnt{\small}
\SetAlCapHSkip{0pt}
\IncMargin{-\parindent}

\acmJournal{TIST} \acmYear{0} \acmVolume{0} \acmNumber{0} \acmArticle{0} \acmMonth{0} \acmPrice{0}
\copyrightyear{2018}

\setcopyright{acmcopyright}

\received{n/a}
\received[revised]{n/a}
\received[accepted]{n/a}

\usepackage{paralist}

\newcommand{\name}{{DeepTracker}\xspace}
\newcommand{\ea}{{$\mathrm{E}_a$}\xspace}
\newcommand{\eb}{{$\mathrm{E}_b$}\xspace}
\newcommand{\ec}{{$\mathrm{E}_c$}\xspace}

\newcommand{\ti}{\textcolor[rgb]{0,0,0}}

\newcommand{\dy}{\textcolor[rgb]{0,0,0}}

\makeatletter
\newcommand\footnoteref[1]{\protected@xdef\@thefnmark{\ref{#1}}\@footnotemark}
\makeatother

\begin{document}
	\title{\name: Visualizing the Training Process of Convolutional Neural Networks}  
	\author{Dongyu Liu}
	\affiliation{%
		\institution{Hong Kong University of Science and Technology}
		\streetaddress{Clear Water Bay}
		\city{Hong Kong}
		\country{China}}
	\author{Weiwei Cui}
	\affiliation{%
		\institution{Microsoft Research Asia}
	}
	\author{Kai Jin}
	\affiliation{%
		\institution{Microsoft Research Asia}
	}
	\author{Yuxiao Guo}
	\affiliation{%
		\institution{Microsoft Research Asia}
	}
	\author{Huamin Qu}
	\affiliation{%
		\institution{Hong Kong University of Science and Technology}
	}
	

\begin{abstract}
Deep convolutional neural networks (CNNs) have achieved remarkable success in various fields. 
However, training an excellent CNN is practically a trial-and-error process that consumes a tremendous amount of time and computer resources.
To accelerate the training process and reduce the number of trials, experts need to understand what has occurred in the training process and why the resulting CNN behaves as such.
However, current popular training platforms, such as TensorFlow, only provide very little and general information, such as training/validation errors, which is far from enough to serve this purpose.
To bridge this gap and help domain experts with their training tasks in a practical environment, we propose a visual analytics system, \name, to facilitate the exploration of the rich dynamics of CNN training processes and to identify the unusual patterns that are hidden behind the huge amount of training log.
Specifically, we combine a hierarchical index mechanism and a set of hierarchical small multiples to help experts explore the entire training log from different levels of detail. We also introduce a novel cube-style visualization to reveal the complex correlations among multiple types of heterogeneous training data including neuron weights, validation images, and training iterations.
Three case studies are conducted to demonstrate how \name provides its users with valuable knowledge in an industry-level CNN training process, namely in our case, training ResNet-50 on the ImageNet dataset. We show that our method can be easily applied to other state-of-the-art "very deep" CNN models.
\end{abstract}

%
%


%
%


%

\maketitle




\section{Introduction}
\label{sec:introduction}

Deep convolutional neural networks (CNNs) have achieved huge success in solving problems related to computer vision, such as image classification~\cite{krizhevsky2012imagenet, simonyan2014very}, object detection~\cite{girshick2014rich}, semantic segmentation~\cite{long2015fully}.
However, in practice, training a high-quality CNN is often a complicated, confusing, and tedious trial-and-error procedure~\cite{bengio2012practical}.
For a large CNN, one complete training trial may take a couple of weeks.
However, domain experts often have to repeat this process several times with slightly different settings to obtain a satisfying network, which may take several weeks or even months.
To accelerate this process, experts have to understand the training processes further to check whether they are on the right track, find latent mistakes, and make proper adjustments in the next trial.
\ti{
Visualizing the concealed rich training dynamics (e.g., the changes of loss/accuracy and weights/gradients/activations over time) is of vital importance to the understanding of CNN training process.
Unfortunately, CNNs usually contain a large number of interacting and non-linear parts~\cite{bengio2013representation} and recently become wider and deeper~\cite{simonyan2014very, szegedy2015going, he2016deep, szegedy2016inception}. Both of these bring considerable difficulties for experts in reasoning about CNN training behaviors.
}

\ti{
Many previous studies investigate what features have been learned by a CNN in one (e.g., usually the last one) or several representative snapshots during the training process~\cite{erhan2009visualizing, zeiler2014visualizing, springenberg2014striving, dosovitskiy2015inverting, mahendran2015understanding, zeng2017cnncomparator, liu2017towards, rauber2017visualizing, alsallakh2018convolutional, Pezzotti2018DeepEyes, kahng2018activis}.
However, little research focuses on visualizing the overall training dynamics.
One recent work~\cite{liu2018analyzing} visualizes the training process of deep neural networks, but it is not tailored for CNNs and not scalable enough to analyze the CNNs that are not only wide but also deep.}
\ti{Besides, tools like TensorBoard, Nvidia Digits, and Deeplearning4j\footnote{TensorBoard: \url{https://www.tensorflow.org/get_started/summaries_and_tensorboard}; Nvidia Digits: \url{https://developer.nvidia.com/digits}; Deeplearning4j: \url{https://deeplearning4j.org/visualization}} are able to show some high-level training dynamics, such as loss and the mean of weights in a layer. However, these tools can neither handle industry-level training (i.e., training a large CNN on a very large dataset) nor answer complex questions. 
For example, how the model's performance on each class of images changes over time? How the changes of parameters impact the classification results for each class? Given so many layers or image classes, which of them are worth paying more attention to and what is the best manner to support comparative analysis?
}
\ti{With these concerns, we are in urgent need of a scalable visualization solution to conducting more advanced analytical tasks.}

To this end, we have to deal with two major challenges.
First, the system needs to handle the large-scale training log data.
Typically, millions of CNN parameters and tens of thousands of validation images are involved in a training process.
In addition, the training is an iteration-based process that usually requires a million iterations to complete, which makes things worse, because the parameters and classification results need to be recorded for every few iterations.
In our experiments (Sec.~\ref{sec:dataprocessing}), a sampled training log may easily exceed a couple of terabytes per training.
To allow an interactive exploration of the data at such scale, it requires not only an effective data storage and index mechanism but also a scalable visualization technique.
Second, the log information is heterogeneous.
The full log contains structure (e.g., neural network), numeric (e.g., neuron weights), image (e.g., validation dataset), and nominal data (e.g., classification results).
Given that significant insights are often hidden underneath the complex relationships among these data, our system also needs to present all these types of data intuitively and assist experts in their analysis tasks.

To address these challenges, we use a downsampling method to store the raw data, and then preprocess and organize these data in a hierarchical manner.
We also design several efficient index mechanisms to support real-time interactions.
To help experts identify the iterations of interest quickly, we propose an application-specific anomaly detection method.
\ti{We also integrate many filtering and aggregation approaches to reduce the amount of presenting information and ensure preserving the noteworthy information.}
For visualization, we design a set of hierarchical small multiples that is combined with a network structure layout to facilitate an effective exploration of the entire training log from different levels of detail.
To reveal the complex correlations among neuron weights, validation images, and training iterations, we further introduce a cube-style visualization.
The cube integrates the small multiples and a matrix-based correlation view, thereby allowing experts to effectively slice and dice the data from different perspectives.


The main contributions of this paper are summarized as follows:
\begin{compactitem}
	\item A systematic characterization of the problem of visualizing the rich dynamics in CNN training processes, and a thorough discussion and summary of the design requirements and space.
	\item A visual analytics system that integrates a tailored large data storage and index mechanism, an anomaly iteration detection algorithm, and a set of well-designed visualization techniques.
	\item A couple of new visualization and interaction techniques including hierarchical small multiples, grid-based correlation view, and cube-style visualization.
\end{compactitem}




\section{Background}
\label{sec:background}


\begin{figure}[htb]
	\centering
	\includegraphics[width=0.8\columnwidth]{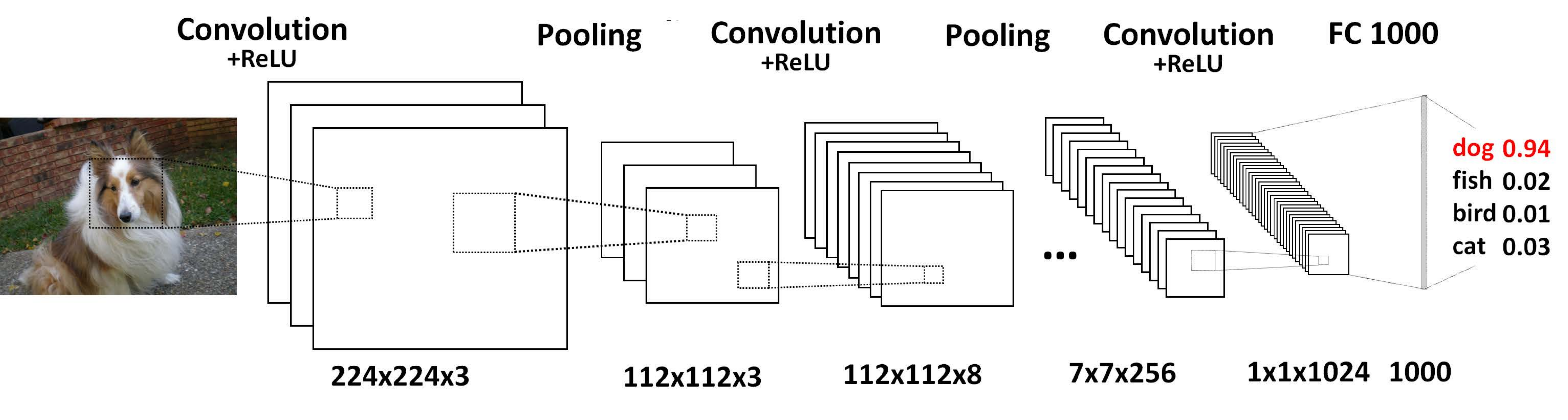}
	\vspace{-4mm}
	\caption{Illustration of a CNN architecture that contains three types of layers (i.e., CONV layer, POOL layer, and FC layer) and transforms an image volume into a class score vector.}
	\label{fig:cnn}
	\vspace{0mm}
\end{figure}

A typical CNN can be viewed as a sequence of layers (Fig.~\ref{fig:cnn}) that transforms an image volume (e.g., a $224\times224$ image with three color channels of R, G, and B) into an output volume (e.g., a vector of size $1,000$ indicating the probability for an input image to belong to $1,000$ predefined classes)~\cite{lecun2015deep}. There are three main types of layers to build a CNN architecture, namely, \textit{convolutional layer} (CONV layer), \textit{pooling layer} (POOL layer), and \textit{fully-connected layer} (FC layer).

A CONV layer comprises numerous neurons that are connected to a local region in the previous layer's output volume through weighted edges, many of which share the same weights through a parameter sharing scheme. 
The weights in each neuron compose a \textit{filter}, the basic unit for detecting  \textit{visual features} in the input image, such as a blotch of color or the shape of an area. 
The output of each neuron is computed via a dot product operation between the weights and inputs from the previous layer, and is then optionally applied via an elementwise activation function (e.g., ReLU, $max(0, x)$). 
A POOL layer is usually inserted between successive CONV layers to reduce the volume of input through a downsampling operation, thereby reducing the amount of parameters and computation in the network. 
The only difference between FC layer and CONV layer is that, in contrast to the neurons in CONV layer that are locally connected and have shared parameters, the neurons in FC layers have full connections to the previous layer's output volume. In addition, the output of the last FC layer is fed to a classifier (e.g., Softmax), computing the scores for all predefined classes, where each score represents the probability that an input image belongs to the corresponding class.


To obtain an effective CNN, the weight parameters in the CONV and FC layers need to be trained using gradient descent methods~\cite{bottou1991stochastic} to ensure the consistency between the predicted class labels and the predefined class for each training image.
Specifically, a training process involves two separate datasets, namely, the training set $\mathit{D_t}$ and the validation set $\mathit{D_v}$. 
To start the training, the parameters of weighted layers are usually initialized via Gaussian distribution~\cite{glorot2010understanding}, and $\mathit{D_t}$ is partitioned into non-overlapping batches.
For each iteration, one batch of images in $\mathit{D_t}$ is fed to the network. 
Afterward, the output classification results are compared with the ground truth (i.e., the known labels of the fed images) to compute the \textit{gradients} with respect to all neurons. 
These gradients are then used to update the weights of each neuron. 
When all batches in $\mathit{D_t}$ are completed (i.e., finish one \textit{epoch}), $D_t$ is reshuffled and partitioned into new non-overlapping batches.
After several epoches, the initially-randomized neural network will be gradually shaped into a specified network targeting at a specific task.
Meanwhile, for every given number of iterations, the network is evaluated via $\mathit{D_v}$.
Similarly, $D_v$ is fed into the network and then the output classification results are collected and compared with the ground truth.
However, the results are only used to validate whether a training process goes well and never used to update neuron weights.



\section{Related Work}
\label{sec:relatedwork}

\subsection{CNN Visualization}
CNNs have recently received considerable attention in the field of visualization~\cite{seifert2017visualizations}. 
Existing approaches can be classified into two categories, namely, feature-oriented and evolution-oriented.

Feature-oriented approaches aim to visualize and interpret how a specific CNN behaves on the input images to disclose what features it has learned.
Most of the existing studies belong to this category.
Some studies~\cite{zeiler2011adaptive,zeiler2014visualizing} modify part of the input and measure the resulting variation in the output or intermediate hidden layers of a CNN. 
By visualizing the resulting variations (e.g., using a heatmap), users can identify which parts of the input image contributes the most to the classification results.
By contrast, other studies~\cite{zeiler2014visualizing,springenberg2014striving,mahendran2015understanding,dosovitskiy2015inverting} attempt to synthesize an image that is most relevant to the activation (i.e., the output of a layer after an activation function) of interest to help experts determine which features of a specified image are important for the relevant neurons.
For example, Mahendran and Vedaldi~\cite{mahendran2015understanding} reconstruct the input images that can either activate the neurons of interest or produce the same activations as another input image.
Besides, some methods focus on retrieving a set of images that can maximally activate a specific neuron~\cite{girshick2014rich,erhan2009visualizing}.
In this manner, users can discover which types of features or images are captured by a specific neuron. 
Built on this method, Liu et al.~\cite{liu2017towards} develop a visual analytics system that integrates a set of visualizations to explore the features learned by neurons and reveal the relationships among them.
In addition, some work~\cite{rauber2017visualizing,chung2016revacnn,kahng2018activis,Pezzotti2018DeepEyes} utilizes dimension reduction technique to project the high-dimension activation vectors of FC or intermediate hidden layers onto a 2D space to facilitate revealing the relationships among outputs.

In contrast to those studies that investigate how a specified network works, only few studies concentrate on visualizing network training processes.
One typical method is to pick several snapshots of a CNN over the training process and then leverage feature-oriented approaches to compare how a CNN behaves differently on a given set of inputs at various iterations~\cite{zeiler2014visualizing, chung2016revacnn, zeng2017cnncomparator, alsallakh2018convolutional}.
For example, Zeiler and Fergus~\cite{zeiler2014visualizing} use deconvnet approach to visualize a series of synthesized images that are most relevant to one activation of interest at a few manually picked snapshots, and then observe the differences between them.
\ti{
Zeng et al.~\cite{zeng2017cnncomparator} present a matrix visualization to show the weight differences of filters of one layer as well as this layer's input and output in two model snapshots. Their system also supports side by side comparison on the learned features of neurons (the computation method is similar to the work~\cite{girshick2014rich}) in different model snapshots.
One limitation of these methods is that when there are myriad filters, images, and iterations, it is challenging to select proper ones to observe and compare.
By contrast, we attempt to reveal the rich dynamics of the network parameters and the classification results at a larger scale to help experts find the notable filters, images, and iterations.
In this point, Alsallakh et al.~\cite{alsallakh2018convolutional} analyze the same data facets (i.e., input images, network parameters, and classification results). However, the work focuses more on identifying the hierarchical similarity structures between classes and still belongs to the category of multiple snapshots comparison, thereby suffering from the same limitation. 
}

To analyze the evolution of CNN parameters, Eliana~\cite{eliana2016pca} treats the parameters of the entire network at one iteration as a high-dimension vector, uses PCA to map the vectors at all iterations onto a 3D space, and creates trajectories for these points. However, this visualization is too abstract to extract useful insights for understanding and debugging the training process.
To analyze the classification results, Rauber et al.~\cite{rauber2017visualizing} create a 2D trails graph to present an overview of how the CNN classification results evolve by leveraging the projection techniques.
However, this method suffers from scalability and visual clutter problems when applied to large-scale datasets (e.g., ImageNet). Besides, this method only provides a very high-level overview and cannot answer those questions that involve individual classes or images.
\ti{The work most similar to ours should be the one by Liu et al.~\cite{liu2018analyzing}, whereas the work mainly targets at deep generative models and would have serious scalability issue if applied in an industry-level CNN training. Besides, compared with that work, we specifically provide a series of hierarchical methods that are tailored for CNNs to visualize the training dynamics, including classification results and network parameters. 
}
\ti{Furthermore, we design a novel correlation matrix and 2.5D cube-style visualization to help experts examine the complex relationships exist among network parameters, classification results, and iterations.}

\subsection{Multiple Time Series Visualization}

Time series data have been extensively studied due to their ubiquity.
Numerous approaches~\cite{aigner2011visualization, aigner2008visual, bach2014review} to time series visualization have been proposed.
Our system also falls into this field, since training logs are essentially a type of time-based information.
Specifically, our system is most related to existing work that visualizes multiple time series.

To visualize multiple time series, one method is to place multiple charts (e.g., line charts) in the same graph space to produce overlapping curves. However, this method reduces the legibility of individual time-series~\cite{javed2010graphical}.
In contrast to overlaying multiple series, one popular alternative is to use small multiples~\cite{tufte1983visual} that split the space into individual graphs, each showing one time series. Using small multiples also allows for an effective comparison across charts. Several factors may also affect the performance of small multiples~\cite{heer2009sizing,javed2010graphical}, such as the types of charts used (e.g., line charts and horizon graphs), the number of series and the vertical space allocated to each series. The improper use of time-series charts, the increased series, and the small vertical space of each series will result in a serious visual clutter problem.

In this work, the evolution of the image classification results of each class and the weight parameters of each layer can be viewed as two types of multiple time series data. Given the large number of classes and layers in a practical CNN training, we identify several space-efficient charts that can characterize the training dynamics. Besides, we propose a similarity-based layout and a hierarchical exploration method to support the exploration of relationships among multiple time series to address the visual clutter problem. We also present a novel cube-based visualization, targeting at the exploration of complex relationships among various types of heterogeneous time-series data (e.g., image classification results, neuron weights, and training iterations).

\section{Requirement Analysis}
\label{sec:requiments}


\name was developed in approximately nine months, in which we collaborate closely with three experts (denoted by \ea, \eb, and \ec) who have considerable experience in CNNs. We held regular discussions with these experts once a week.


From our regular discussions, we learned that the training process should be carefully babysat. The experts have to tune and fix a number of \textit{hyper-parameters} (learning rate, batch size, layer number, filter number per layer, weight decay, momentum, etc.) before starting a training trial.
These hyper-parameters, which strongly influence how a CNN is trained, are usually selected based on the experiences learned from their previous trials. 
During a training, there are several useful quantities the experts want to monitor, such as loss function (the residual error between the prediction results and the ground truth), train/validation error rate (the percentage of mislabeled images), weight update ratio (the ratio of the update magnitudes to the value magnitudes), and weight/gradient/activation distributions per layer. Basically, both loss and error rate should decrease over time; the consistent increase or violent fluctuation of loss on $D_t$ may indicate a problem; a big gap between the error rates of $D_t$ and $D_v$ may suggest that the model is over-fitting, while the absence of any gap may indicate that the model has a limited learning capability; the update ratio\footnote{In most cases, if the update ratio is lower than 1e-3, the learning rate might be too low; if it is higher than 1e-3, the learning rate is probably too high.} is expected to be around 1e-3; the weight distributions per layer in the beginning should overall follow Gaussian distributions with different standard deviation settings\footnote{\label{note7}Xavier Initialization~\cite{glorot2010understanding} is applied in our model. Basically, the deeper (close to loss layer) the layer is, the smaller the sd is.} and may become diverse after training; and exploding or vanishing gradient values are a bad sign for the training.

The aforementioned rules of thumb are all based on the analysis of high level statistical information.
However, the experts still strongly desire to examine more details underlying the statistic, so that they can gain more knowledge and suit the remedy to the case when problems occur. 	
For example, the experience tells the experts that it is better to continue training models (e.g., ResNet~\cite{he2016deep}) with the same learning rate for a while rather than turn it down immediately, when the overall error rate stops to decrease. The experts are very curious about what happens to the model and its performance on each class of images during the period. 
Also, we may often see a fluctuation of loss or an occasional exploding gradient values, and what brings about this? Are there any layers or filters that behave abnormally? 
These details are never uncovered before.
Thus, the experts strongly desire a tool to enable them to explore the hidden dynamics behind a CNN training. 
After several rounds of structured interviews with the experts, we finally summarized and compiled the following list of requirements:

\begin{compactenum}[R.1]
	
	\item \textbf{Exploring multiple facets of neuron weight information}.
	A single iteration may have millions of weight updates, but individual values are generally meaningless to the experts.
	Instead, the experts are more interested in the statistical information of these weights at different levels of detail (e.g., CONV layer level and filter level).
	All three experts emphasized the importance of intuitively and effectively examining general types of statistics, such as sum and variance, between these levels.
	Besides, \ea and \eb strongly desire to see the weight change degree for filters over iterations to identify which filters change dramatically at which iteration or how a filter changes across iterations.

		
	\item \textbf{Comparing multiple layers}.
	All three experts like to compare layer-level statistical information. 
	For example, they want to know whether a specified measure of different layers show a similar trend or distribution (e.g., whether the sum is positive or negative).
	Accordingly, our visualization should also help these experts in performing such comparisons. 
		
	\item \textbf{Tracking the classification results of validation classes}.
	Validation is a critical step in the training process that ``test drives'' the trained CNN and tracks its performance change across iterations~\cite{bengio2012practical}.
	Previous tools only measure the global validation loss/error, thereby concealing the rich dynamics of how the image labels of each class change over time.
	When the error rates do not reduce as expected, the experts find that such highly-aggregated information is useless and cannot help them understand why or when the training runs into a bottleneck.
	Therefore, the experts want to know how the images of different classes behave differently and identify those classes that are easy or difficult to train.

	\item \textbf{Detecting important iterations}.
	One complete training process usually contains millions of iterations, but it is obvious that not all iterations are equally important.	
	For example, some image classes may suddenly change their error rates after certain weight updates.
	The experts are interested in these patterns, which may help reveal in-depth relationships between filters and image features.
	However, the overall error rate trend does not help much, since it generally decreases slowly and steadily.
	Thus, the experts hope that our system can help them identify the abnormal iterations and the corresponding classes.
	
	
	\item \textbf{Examining individual validation classes}.
	Our initial prototype shows that different classes clearly have different error rate evolution patterns. 
	Thus, experts \eb and \ec are curious about those classes with very poor or excellent performance, thereby desiring to further explore the image label information for these classes. 
	For example, they want to see whether and why some images are always misclassified. 
	
	\item \textbf{Enabling correlation exploration}.
	Apart from analyzing the weight and validation data separately, the experts are also curious about their relational patterns.
	They are specifically interested in uncovering the relationships between the layers or filters and the validation images, such as how the changes in network parameters respond to the image labeling results for each class. 
	By connecting these two pieces of information together, they hope to gain fundamental insights into the behaviors of CNNs and to improve training results.
\end{compactenum}


\section{System Overview}
\label{sec:systemoverview}

\begin{figure}[htp]
	\centering
	\includegraphics[width=0.8\columnwidth]{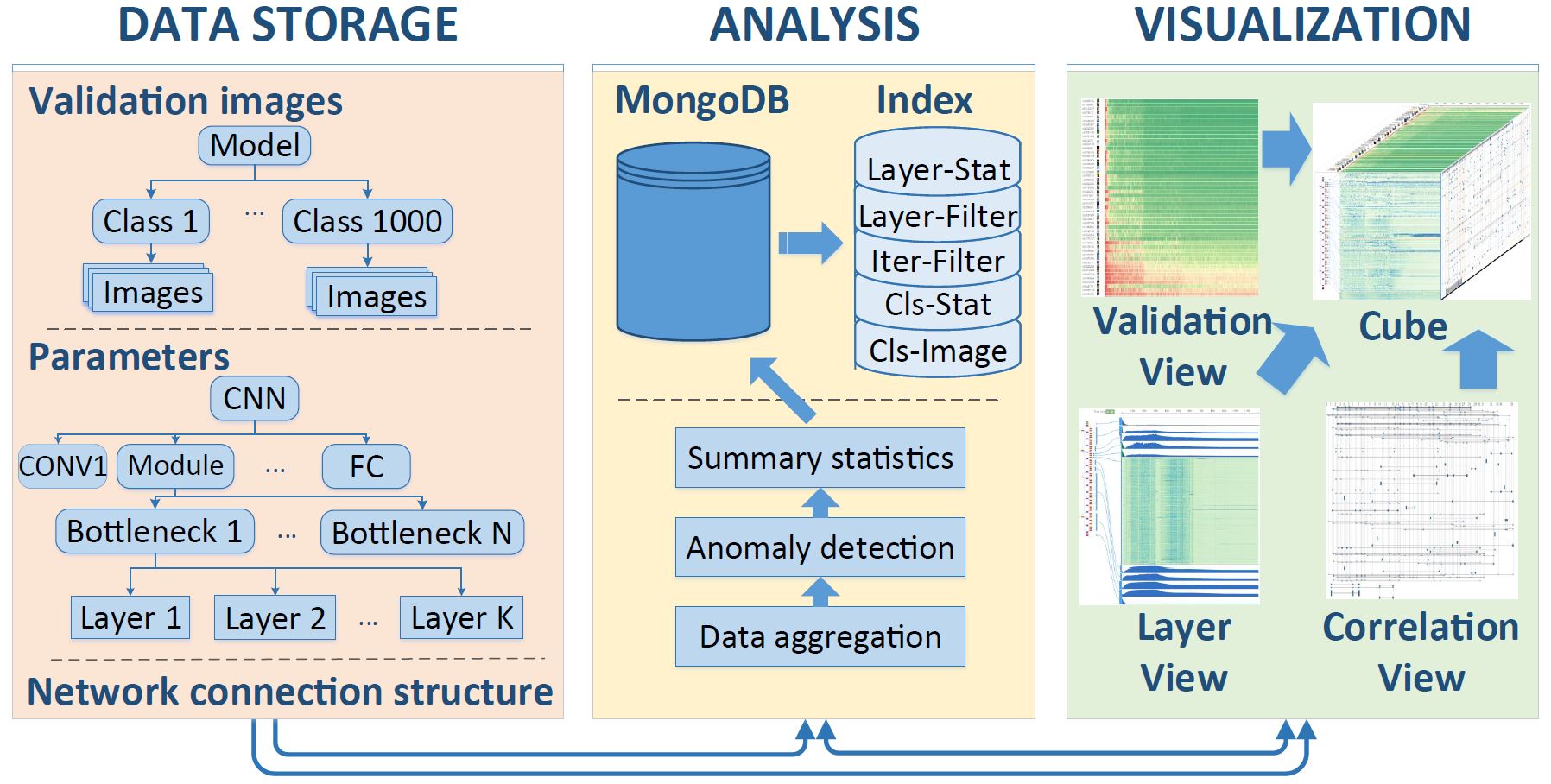}
	\vspace{0mm}
	\caption{Three components of \name. The raw data are preprocessed into a hierarchical structure and then stored into five application-specific data indexes to enable real-time interactions. On top of the efficient data storage, several visualizations are combined together to help experts with analysis tasks from different levels and perspectives.}
	\label{fig:system_overview}
	\vspace{0mm}
\end{figure}

\name is a web-based application developed under the full-stack framework, MEAN.ts (i.e., MongoDB, Express, AngularJs, Node, and Typescript). The back-end part of our application is deployed in a server with 3.10GHz Intel Xeon E5-2687W CPU and 32GB memory. 

The architecture of our system (Fig.~\ref{fig:system_overview}) begins with the data processing part, where the entire training log is hierarchically organized, and several application-specific indexes are built to support real-time interactions.
On top of the efficient data storage, we build three coordinated views, namely, the Validation View, the Layer View, and the Correlation View, to support an interactive analysis from different levels and perspectives.
The Validation View aims at providing a visual summary of CNN performance on different validation classes. 
By combining our anomaly detection algorithm and small multiples, experts can easily identify different image class behavior patterns and critical iterations for each image class (\textbf{R3}, \textbf{R4}). 
Experts may also drill down to the class of interest to explore further image label information (\textbf{R5}).  
The Layer View aligns the weight information with the CNN structure to help experts explore various statistical information in the network hierarchy. 
Experts can further drill up or down in the network hierarchy to compare these measures at different levels of detail (\textbf{R1}, \textbf{R2}).
The Correlation View presents a novel grid-based visualization method to provide an overview of the correlation between the image classification results and the neuron weights of each filter (\textbf{R6}).
\ti{
The three views compose a cube, with which the experts can simultaneously explore the changes of class-level performances, the variations of filter-level weights, and the correlations between them.
}

\section{Data Acquisition and Construction}
\label{sec:dataprocessing}

\ti{
The primary motivation of this work is to monitor industry-level CNN training processes. Therefore, we conduct our experiments with ResNet-50~\cite{he2016deep} and ImageNet Dataset~\cite{russakovsky2015imagenet}.
ResNet-50, containing 50 weighted layers (i.e., CONV and FC layers), is among the most popular CNNs that have been recently used in practice and meanwhile ImageNet 2012 is also among the largest and most challenging publicly available datasets.  
The dataset includes $1,000$ classes of images, each class containing $1,300$ training images and $50$ validation images.
Training such a model needs around 120 epoches (nearly 1.2 millions iterations when batch size is 128) to achieve convergence.
Simply dumping all the information for every iteration can easily have the size of dumped data exceed several petabytes and take about 4 weeks. Through discussion, we all thought that $1,600$ is a reasonable interval to capture meaningful changes (about 7 times per epoch). This reduces the log to a manageable size (about 1TB).
}

\ti{
For each dump, we recorded two pieces of information, namely, neuron weights/gradients of CONV layer and FC layer and image classification results. 
The parameters on BN layers were not recorded, as they can be totally recovered given the weights on CONV and FC layers and always need to be updated when applied in a new dataset.
Besides, we did not record the activations of each layer/filter for every validation image, as doing so is technically impracticable considering the extremely large models and datasets and the limited disk storage. Further, activation evolution visualization is beyond the research scope of this paper. 
Sec.~\ref{sec:conclusion} discusses activation visualization is indeed a perfect complementary technique to our work.
}
We organized the weight/gradient information according to the natural hierarchial structure of ResNet-50. It consists of four \textit{CONV modules} (plus the first CONV layer and the final FC layer, there are 50 layers in total). Each module contains several \textit{bottleneck blocks}~\cite{he2016deep} that comprise three to four basic CONV layers (data storage in Fig.~\ref{fig:system_overview}).
Thus, we grouped all neuron weights to align with such hierarchy.
\ti{
In a similar manner, we organized the classification results hierarchically from individual level, class level, to model level.
We stored all the data into MongoDB\footnote{MongoDB is a free and open-source cross-platform document-oriented (NoSql) database. \url{www.mongodb.com}}.
In particular, we precomputed all the relevant aggregation values, such as weight means and error rates, for each filter, layer, image, and class.
Nevertheless, the distilled data still remain too large to load into memories (about dozens of gigabytes per training).
Therefore, we analyzed the frequent needs of the experts and built several indexes to enable real-time interactions, including Layer-Stat index $I_{ls}$, Layer-Filter index $I_{lf}$, Iter-Filter index $I_{if}$, Cls-Stat index $I_{cs}$, and Cls-Image index $I_{ci}$.
\begin{compactitem}
	\item $I_{ls}$ retrieves the statistic values (e.g., mean and sd) at every iterations for any given layer;
	\item $I_{lf}$ lists all the filter-level information (e.g., changing degree of each filter) at every iterations for any given layer;
	\item $I_{if}$ searches the top changing filters from all layers at any given iteration;
	\item $I_{cs}$ extracts class-level information (e.g., class performances, the different types of abnormal images~Sec.~\ref{sec:anomaly}) over all iterations for any given class;
	\item $I_{ci}$ fetches the meta-data of images for any given classes.
\end{compactitem}
}

\section{Visualization}
\label{sec:visualization}

In this section, we describe our three coordinate views, namely, the Validation View, the Layer View, and the Correlation View, that help experts accomplish the aforementioned analytical tasks.

\subsection{Validation View}

\begin{figure}[htbp]
	\centering
	\includegraphics[width=0.95\columnwidth]{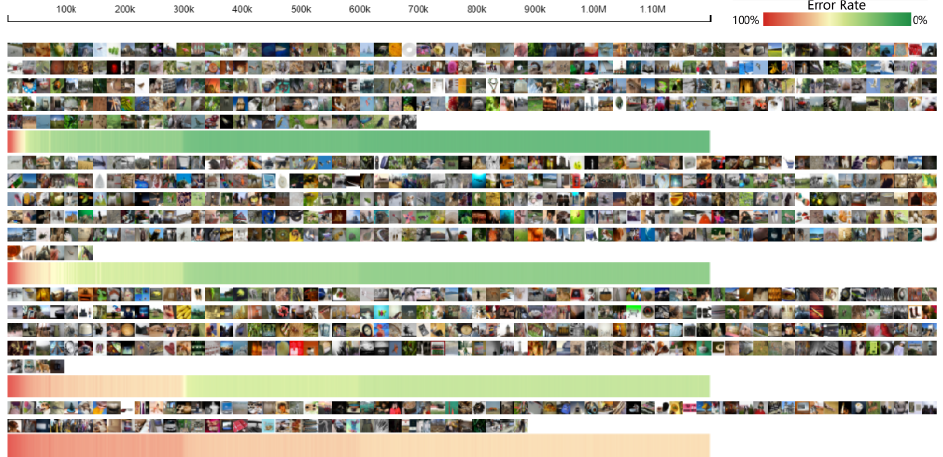}
	\vspace{-2mm}
	\caption{From top (a) to bottom (b), image classes are more and more difficult to train.}
	\label{fig:case1-2}
	\vspace{-4mm}
\end{figure}

Several urgent requirements (\textbf{R3}, \textbf{R4}, \textbf{R5}) from the experts need to examine how the evolving CNN acts differently on the validation images of each class rather than how the overall validation error rate differs over training.
Thus, we design the Validation View (Fig.~\ref{fig:case1-2} \& \ref{fig:case1}) to present all image classes in $D_v$.

\subsubsection{Visual Encoding}
\label{sec:validation_view_encode}

\ti{By default, the view starts with a visualization of \textbf{cluster-level} performance (\textbf{R3}).
The classes with similar evolving trends form a cluster and then their error rates at every iteration are averaged.
The averaged error rates are then depicted as a colored stripe, where the x-axis encodes the iterations and the error rates are encoded by colors (Fig.~\ref{fig:case1-2}).
We choose k-means as the clustering algorithm and $k$ can be adjusted according to demands (Fig.~\ref{fig:case1-2} shows the case when $k=4$).
Experts can open up one cluster to further examine the performance in \textbf{class-level} (\textbf{R3}). 
The design is based on the following considerations.
First, the experts are more interested in the overall classes than in individual iterations.
Thus, small multiples technique (juxtaposed techniques) is chosen for their superior performance in high-level comparison tasks (e.g., global trends for every series)~\cite{javed2010graphical}.
Second, we cannot present all the classes ($1,000$ in ImageNet) at one time, we have to consider a hierarchical and highly space-efficient visualization.
However, many traditional charts, such as line charts and horizon graphs, require a larger vertical space~\cite{heer2009sizing} than 1D heatmaps.
Compared with traditional charts, heatmaps are also more easy to do side by side comparison for their symmetrical space (i.e., no irregular white spaces).
As a result, all experts prefer the heatmap-based small multiples.}

\textbf{Image-level performances, R5}. The class-level color strips can be further unfolded to explore the image-level evolution patterns.
Unfolding a heatmap reveals a pixel chart (Fig.~\ref{fig:case1}d), with each row (1px height) representing an image and each column (1px width) representing an dumped iteration (consistent with the class heatmap).
We use red and green colors to indicate the incorrect and correct classifications, respectively.
Meanwhile, the experts can zoom/pan the pixel chart for a closer inspection. Clicking on a row shows the original corresponding image.

\textbf{Anomaly iterations, R4}. \ti{As experts are concerned about the iterations with abnormal behaviors, we particularly propose a algorithm to detect these anomaly iterations (refer to Sec.~\ref{sec:anomaly}).
Experts can choose to only show the classes with anomaly iterations (Fig.~\ref{fig:case1}).
At this point, for each class-level color stripe, we use triangular glyphs to highlight these detected anomaly iterations.
The upside-down triangles ($\bigtriangledown$) and normal triangles ($\triangle$) indicate those anomaly iterations that are detected by the left-rule and right-rule, respectively.
The widths of triangles encode the anomaly scores. 
Experts can set a threshold value to filter the triangles with low anomaly scores.}

\subsubsection{Anomaly Detection}
\label{sec:anomaly}

In our scenario, the classification results for an image can be represented by a 0/1 sequence ($[a_1, ..., a_n]$), where each element represents a correct or incorrect result at the corresponding validation iteration.
The experts are curious about the iterations when a significant amount of 1/0 flips (i.e., 0 to 1 or 1 to 0) occur for a class.
In general, this problem can be modeled and solved using Markovian-based anomaly detection algorithms~\cite{aggarwal2015outlier}.
Despite the popularity of using Markovian methods to detect outliers in discrete sequence, we decide to employ rule-based models~\cite{aggarwal2015outlier} for two reasons.
First, Markovian methods are a black box and the resulting outlier values are sometimes difficult to comprehend. Second, the experts have explicitly described two types of iterations they are very interested in, namely, those iterations when many images with values that remain stable for many previous iterations suddenly flip (denoted by the \textit{left-rule}) and those iterations when many images flip and keep their values stable after many iterations (denoted by the \textit{right-rule}).
Fortunately, these anomalies can be easily modeled using rules.
The rule-based models primarily estimate the value $P(a_i|a_{i-k},\ldots,a_{i-1})$, which can be expressed in the following rule form: $ a_{i-k},\ldots,a_{i-1} \Rightarrow a_{i}.$
In our scenario, if an image has the same value (either 0 or 1) in the previous consecutive $k$ iterations ($i-k,\ldots, i-1$), then its value must be the same at iteration $i$ (the left-rule).
Otherwise, iteration $i$ is considered an outlier for the specified image.
Based on these considerations, we develop an application-specific algorithm to detect anomaly iterations in the validation history. The algorithm includes the following steps:
\begin{compactenum}
\item{\textbf{Rule-Judgement}:}~The algorithm computes a vector $[l_{i1}, \ldots, l_{in}]$ for every image $i$, where $l_{ij} =1$ if the left-rule is satisfied, otherwise, $l_{ij} =0$;

\item{\textbf{Aggregation}:}~For each class that contains $m$ images, the algorithm aggregates all the computed vectors for each image into one $[L_1, ..., L_n]$, where $L_j = \sum_{i=1}^{m}l_{ij}$, \ti{denoting the left anomaly score at iteration $j$ for this class.}
\end{compactenum}

The approach is a window-based method, and the experts can adjust the window size $k$ to control the sensitivity of the anomalies.
In a similar manner, we detect the anomalies from the opposite direction for the right-rule.


\subsection{Layer View}

\begin{figure}[htbp]
	\centering
	\includegraphics[width=0.95\columnwidth]{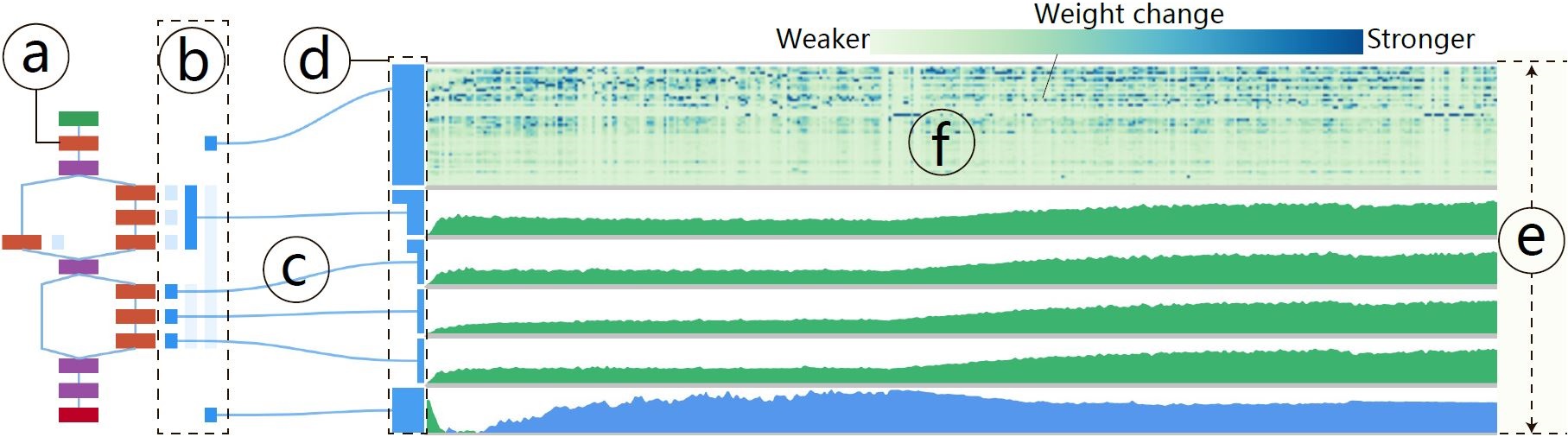}
	\vspace{-2mm}
	\caption{Visual encodings in the Layer View: (a) a CONV layer, (b,d) hierarchical bars, (c) links between the CNN structure and (e) the hierarchical small multiple charts, (f) a pixel chart for one layer.}
	\label{fig:layer_view_illustrator}
\end{figure}

The Layer View focuses on weight-related tasks (\textbf{R1}, \textbf{R2}).
The view consists of two connected parts, namely, the CNN structure and the hierarchical small multiples (Fig.~\ref{fig:layer_view_illustrator}), so that experts can hierarchically explore and compare various types of statistic in the context of the network structure.

\textbf{CNN structure}.
The experts hope our tool can help them explore the statistical information of each layer and meanwhile know their relative positions in the entire network (\textbf{R1}).
Thus, we adopt Netscope\footnote{Netscope is a web-based tool for visualizing neural network architectures. \url{http://ethereon.github.io/netscope/}}, a popular neural network visualizer, in our system.
\ti{
The green rectangle is data input layer, the red ones are the CONV layers, and the purple ones mean the pooling layers.
The links between these rectangles show the network structure.
We further add blue level bars (Fig.~\ref{fig:layer_view_illustrator}b) to encode the latent hierarchy (from CONV modules, bottlenecks to basic CONV layers, Sec.~\ref{sec:dataprocessing}).
}
The right most level bars represent CONV modules (Sec.~\ref{sec:systemoverview}), which are recursively divided into smaller level bars until reaching elementary CONV layers (i.e., red rectangles).

\textbf{Hierarchical small multiples}.
To assist experts in exploring and comparing layers of a deep CNN (\textbf{R1, R2}), a space-efficient visualization technique is demanded.
Thus, we leverage hierarchical small multiples to show layers of interest (Fig.~\ref{fig:layer_view_illustrator}e).
\ti{By default, experts are presented with the information about CONV modules and then can drill down to see more information about low-level CONV layers with interactions with the network graph (i.e., click on the corresponding level bars).
The width of outcropping rectangles (Fig.~\ref{fig:layer_view_illustrator}d) encodes the aggregation level of current layer charts. For example, the top second layer chart in (Fig.~\ref{fig:layer_view_illustrator}e) shows the bottleneck-level aggregation information and the following three layer charts show the basic CONV layer level information. Besides, the links (Fig.~\ref{fig:layer_view_illustrator}c) mark the real positions of the layer charts in the network structure.
}

The small multiples support multiple types of charts including line chart, horizon graphs~\cite{heer2010tour} and box plots to emphasize the different aspects of the statistical data.
The experts use box-plots to see the rough distribution of statistical values and use basic line charts to examine individual values. Besides, the experts prefer to use horizon graphs when performing tasks in regard to trend tracking and comparison (\textbf{R2}), because of its effectiveness in visualizing divergent weight values~\cite{javed2010graphical}.
Similar to unfolding the class heatmap to a pixel chart, the experts are also allowed to open the layers of interest as a pixel chart (Fig.~\ref{fig:layer_view_illustrator}f) that presents the filter-level information (\textbf{R1}). Each row (1px height) in the pixel chart represents one filter, and each column (1px width) indicates one iteration. We use sequential colors to encode pixel value (e.g., the cosine similarity between two subsequent dumped iterations).

\subsection{Correlation View}
\label{sec:correlation}


This view  helps experts establish connections between the filters and images.
In particular, the experts want to understand further how the changes in network parameters are related to class performances (\textbf{R6}).
For example, several anomaly iterations may be detected for a single class.
For each detected anomaly iteration, we can identify a set of \textit{anomaly filters} (i.e., the top $k$ filters with largest changes at that iteration).
Since different classes may share anomaly iterations and different anomaly iterations may share anomaly filters, are there any filters that are commonly seen in these iterations?
Do any of the anomaly classes or filters strongly co-occur?
We designed the Correlation View to answer these questions.


\begin{algorithm}[t]
	\SetAlgoNoLine
	\KwIn{Target set $S_{target}$.}
	\KwOut{Minimum partition for $S_{target}$.}
	$S_{result}$ = $\emptyset$\;
	\For{each target set $s_t$ in $S_{target}$}{
		$S_{new}$ = $\emptyset$\;
		\For{each mini set $s_r$ in $S_{result}$}{
			$itersection$ = $s_t \bigcap s_r$\;
			$S_{new}$ = $S_{new} \bigcup intersection \bigcup (s_r - s_t)$   \label{alg:setsplit:split}\;
			$s_t$ = $s_t - s_r$\;
		}
		\If{$s_t$ is not empty}{
			 $S_{new} \bigcup s_t$\;    \label{alg:set:addremain}
		}
		remove all empty set in $S_{new}$\;
		$S_{result}$ = $S_{new}$\;
	}
	{\bf return} $S_{result}$ \label{alg:setsplit:return}

\caption{Minimum Set Partition}
\label{alg:setsplit}
\end{algorithm}

\textbf{Filter set partition}.
We first introduce the \textit{mini-set} concept to organize anomaly filters that are shared by multiple anomaly iterations and different classes.
For each class $C_i\in\{C_i|{1\leq i \leq n}\}$, we denote its anomaly iterations by $T_i = \{t_{i,k}|{1\leq k\leq n_i}\}$.
Thus, all anomaly iterations are $\cup_{1\leq i \leq n}T_i$, denoted by $T$.
For each anomaly iteration $t\in T$, we denote its anomaly filters at CNN layer $L_j\in\{L_j|{1\leq j \leq m}\}$ by $s_{j,t}$.
Thus, for each layer $L_j$, we can collect all anomaly filter sets $\{s_{j,t}|t\in T\}$ (denoted by $S_j$)  and all anomaly filters $\cup_{t\in T}s_{j,t}$ (denoted by $s_{j}$).
Thus, mini-set aims to find a minimum set partitions of $s_{j}$ (denoted by $s_j^\ast$) that each $s_{j,t}$ can be assembled from some elements (i.e., mini-sets) in $s_j^\ast$.
We specifically propose a Set Partition Algorithm (Alg.~\ref{alg:setsplit}) to find $s_j^\ast$.
The algorithm accepts a target set as input (i.e., $S_j$). $S_{result}$ is initially empty and a new anomaly filter set is used at each time to partition the mini-sets contained in $S_{result}$ (cf. lines~4 to 7). If the new anomaly filter set is not empty after partitioning, then it is added as a new mini-set (cf. line~8). Finally, the partitions contained in $S_{result}$ will be returned.

\begin{figure}[t]
	\centering
	\includegraphics[width=1\columnwidth]{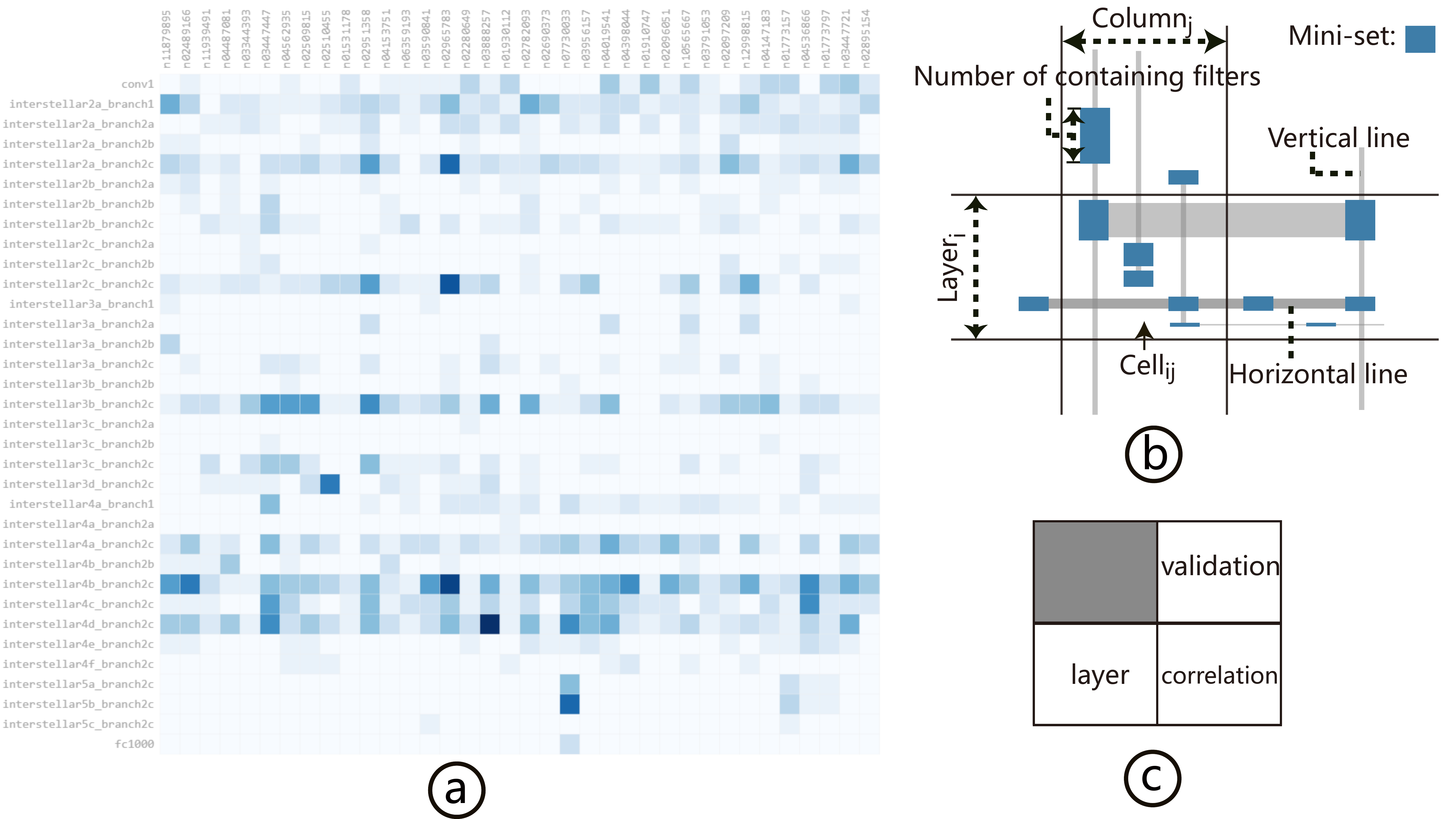}
	\caption{ (a) The abstract version of correlation view, where rows and columns represent layers and image classes, respectively. A sequential color scheme is used to encode the number of anomaly filters. (b) The complex version of correlation view, where the detailed information of individual anomaly filters are shown. (c) A layout solution for coordinated analysis without using skewed axes. }
	\label{fig:correlation_view}
\end{figure}



\textbf{Visual encoding.}
To intuitively represent these relationships, we introduce a grid-style visualization (Fig.~\ref{fig:correlation_view}), where rows and columns represent layers and image classes, respectively. The number of rows and columns equal to the number of layers with anomaly filters and classes with anomaly iterations, respectively.
\ti{
We start from a abstract version.
In this version (Fig.~\ref{fig:correlation_view}a), for $\mathrm{Cell}_{i,j}$, a sequential color scheme is used to encode the number of anomaly filters ($\cup_{t\in T_j}s_{i,t}$). 
The darker the color it is, the more anomaly filters appear in layer $L_i$ that are related to class $C_i$.
From this visualization, we can easily observe the correlations between layers and classes, while it also hides much detailed information.
We cannot answer questions like whether the filters in $\mathrm{Cell}_{i,j}$ are the same with $\mathrm{Cell}_{i,k}$ (this kind of information shows how many classes this filter can impact and help examine the relationships among classes), or whether there are filters in $\mathrm{Cell}_{i,j}$ appearing in more than one anomaly iteration (this shows the importance of these filters for class $C_j$).
}

To solve these problems, we provide an advanced version (Fig.~\ref{fig:correlation_view}b).
For $\mathrm{Cell}_{i,j}$, the width and height encode the number of anomaly iterations and the number of anomaly filters of the corresponding class and layer (i.e., $|T_j|$ and $|s_{i}|$), respectively.
Based on these numbers, the columns and rows are further divided with vertical/horizontal lines.
For a class (e.g., $\mathrm{Column}_{j}$), $|T_j|$ vertical lines are drawn to represent all related anomaly iterations (i.e., $[t_{j,1}, t_{j,2},\ldots,t_{j,n_j}]$).
For each row of layer $L_i$, there are  $|s_i^\ast|$ horizontal lines representing all mini-sets in that layer.
The intersections between these horizontal and vertical lines are highlighted with blue rectangles if the corresponding mini-set is part of the anomaly filters of the corresponding anomaly iteration.
The height of the rectangles represents the number of filters of the corresponding mini-set.
\ti{Obviously the introduction of mini-sets dramatically cuts down the number of horizontal lines and blue rectangles, otherwise each anomaly filter require one horizontal line and one rectangles, which may cause serious visual clutter problem. In fact, mini-sets can be viewed as a partially aggregation version instead of representing all the anomaly filters as horizontal lines and rectangles.
Users can set the minimum appearing number of mini-sets to filter the sets with lower importance.}

\subsection{Cube Visualization}
\label{sec:cubeview}


The log data contain three main aspects of information, namely, iterations, validation information, and weight information.
The three aforementioned views are designed to show all possible 2-combinations of these three types of information, respectively.
\ti{
Although these views can be used individually, they need to be combined together to form a complete picture.
}
Thus, we propose a novel and intuitive visualization technique, which naturally and seamlessly stitches the three views together, based on their shared axis (inspired from Binx~\cite{berry2004binx}), into a ``cube'' shape (Fig.~\ref{fig:teaser}).
When experts find or highlight a pattern of interest, they can easily track the pattern over the edges to find the related information in the other two views easily. 

\ti{
The use of skewed axes may bring about a possible perspective distortion problem. Nevertheless, the advantages of the cube-style design far outweigh the disadvantages.
Given the limited pixels in a computer screen and each view requiring a large display space, the experts all agree that the cube-style design is the most space-efficient and intuitive manner to show all the information. Furthermore, with such design, it prevents the experts from switching from multiple views, reducing the cognitive burden and the load of memory.
This allows the experts to conduct correlation analysis more effectively.
We also provide a compromised solution to handle the distortion problem, that is, laying out the three views in the form like Fig.~\ref{fig:correlation_view}c. So the experts can firstly examine the layer view (horizontally) or validation view (vertically) together with the correlation view, and then switch to cube mode to explore the three views together.
}

\ti{
Notice that there are some different settings for several views in the cube.
For the layer view (front), only the layers with anomaly filters are activated (see the activated blue bars in the front view of  Fig.~\ref{fig:teaser}), the weight variation of each anomaly filter is represented as a horizontal color strip.
For the validation view (top), only the classes with anomaly iterations are preserved.
The following lists several common exploration pipelines:
\begin{compactitem}
	\item P1: from the layer view (front), we can quickly check the distribution of activated layers in the overall network and pick some anomaly filters of interests. Then, by tracking along the horizon axis to the correlation view (right), we can examine which classes these filters impact and how important these filters are to the classes. Finally, we can observe the evolving patterns of these classes and the corresponding anomaly iterations in validation view (top).
	\item P2: from the validation view (top), we can firstly mark several anomaly iterations of some classes. Then, we can check the corresponding columns in the correlation view, finding the rows that contain anomaly filters and exploring the importance of these filters to these classes and how these filters impact the other classes. Finally, by highlighting these corresponding rows, we can observe them in the layer view to see how these filters behave around the picked anomaly iterations.
	\item P3: from the correlation view (right), we can search the horizontal lines across many rectangles (it means these filters impact many classes at the same time) or the rectangles that appear more than one time in the same cell (these filters are judged as anomalies many times for a class and may have great impact on this class). With these selected horizontal lines or rectangles, we can simultaneously track their corresponding weight variation information in the layer view (front) and class performance information in the validation view (top).
\end{compactitem}
}

\section{Use Examples}
\label{sec:evaluation}

We derived these examples through the assistance of our collaborating experts, who were familiar with our designs and data. As a remark, the following results are from the experiment with 8 times larger batch size and learning rate setting than the basic setting introduced in Sec.~\ref{sec:dataprocessing}.

\subsection{Exploring Validation Results}

\begin{figure}[htp]
	\centering
	\includegraphics[width=0.8\columnwidth]{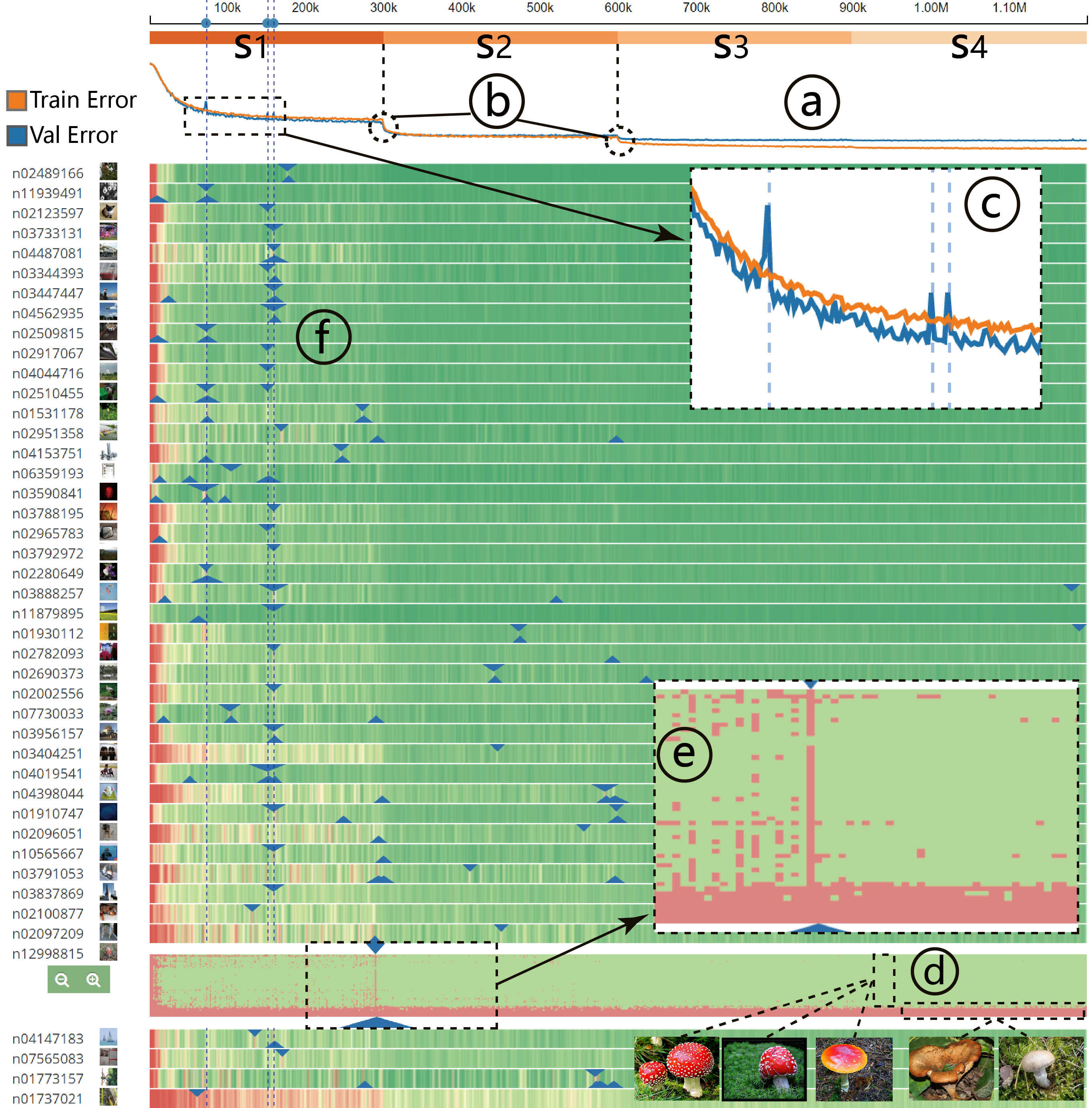}
	\vspace{-2mm}
	\caption{Overview of validation classes. (a) Two curves show the overall training/validation error rate. (b) Two turning points align well the boundary of stage $s2$. (c) Three peaks appear in the stage $s1$ and align well with (f) the detected anomaly iterations. (d) Two types of mushroom images have different behaviors in the class. (e) Most images in the class flip at the anomaly iteration.}
	\label{fig:case1}
\end{figure}

The first scenario demonstrates how the experts use \name to explore the image classification results (\textbf{R3}, \textbf{R4}, and \textbf{R5}).


\textbf{Performance evolution patterns}.
Fig.~\ref{fig:case1}a shows a typical visualization of training/validation errors that may appear on any popular training platform. The timeline at the top shows a total of 1.2 million iterations.
Beneath the timeline, four line segments represent four stages ($s1$, $s2$, $s3$, and $s4$ in Fig.~\ref{fig:case1}) in the training process, \dy{where the later stage has one-tenth of the \textit{learning rate} of the previous stage.} 
We can observe that two sudden drops in the curves match well with the boundaries of the training stages (Fig.~\ref{fig:case1}b).
However, this is a well-known pattern to the experts.
On the other hand, although the overall error rate continues to decrease, the class-level error rates show a more complicated story, that is new to the experts.
By quickly scanning the small multiples in cluster-level, the experts identify there are generally four types of class evolving patterns (Fig.~\ref{fig:case1-2}).
From top to bottom, the four types are more and more difficult to train.
\ti{For example, for the type at the top, these classes are recognized correctly after a few iterations.
By contrast, the classes at the bottom always have high error rates in the entire training process, which means that the resulting network fails to recognize the related images.}
From this, the experts learn that the model has spent most of time to improve its performance on the classes of middle-level classification, since the model has already performed well on the easy-trained classes at a very early stage and is always performing miserably on the hard-trained classes over the entire training. 
From these patterns, the experts consider it promising to accelerate the training process and improve the overall performance by treating classes differently during the training process. That is, stop feeding the easy-trained classes in an appropriate early stage, put more efforts on training the classes of middle difficulties for classification, and figure out why some classes always have extremely high error rates. One similar attempt has been made in a recent work~\cite{lin2017focal}.

\textbf{Anomaly iterations}.
The experts are curious about the three sudden peaks in stage $s1$, and then mark these three iterations with dotted lines (Fig.~\ref{fig:case1}c), which look like anomaly iterations (\textbf{R4}).
However, the colors in the small multiples do not have clear patterns related to these iterations.
Then, the experts turn on the anomaly detection and immediately find that many triangles are aligned well with the dotted lines (Fig.~\ref{fig:case1}f), thereby confirming our suspicion.
Then, the experts can click on the corresponding image icons to see the detailed images that contribute to the three peaks.
In addition, there are more anomaly iterations in stage $s1$ then in the later stages.
This interesting pattern can be explained by the reduction of learning rate and the convergence of the model in the later stages. At the same time, it also implies that the learning rate in stage $s1$ is slightly too high, leading to the instability in the model (the case in Sec.~\ref{sec:case2} indicates the same finding for the discovery of potential ``dead'' filters).


\textbf{Details in classes}.
To further examine what happens at the anomaly iterations for a class, the experts can further check the image-level information of the class (\textbf{R5}).
For example, the experts are curious about the abnormally large anomaly iteration in the class of ``mushroom'' (Fig.~\ref{fig:case1}d) that are captured by both the left-rule and the right-rule.
Then, they click and expand color stripe to see the pixel chart of images.
First, they confirm that this iteration is indeed special for this class, because nearly all images flip during particular that \dy{dumped interval} (Fig.~\ref{fig:case1}e).
Thus, they may further investigate to find the layers or filters that cause such flips based on the filter updates around that iteration.
\dy{In particular, the experts comment that, it seems that after the iteration, the CNN model has jumped to a better local optimal for the class, because the green color is more stable after the iteration.
This may result from the reduction of the learning rate (from $s1$ to $s2$).  
This kind of patterns appear frequently in many classes during the whole training process, many of them occurring not in the learning rate transition point.
The experts wonder that the model should be trying to jump from one local optimal to another better local for these classes continuously, so as to reduce the overall error rate gradually.}
This insight has never been obtained because the experts initially thought that the error rate for one class should decrease steadily.
Besides, the experts also find that, at the bottom of the pixel chart, several images are mislabeled in the entire training process, although the class is easy to train overall(Fig.~\ref{fig:case1}d).
To understand why, the experts click on these images to examine them, and find that the contents in the mislabeled images have a clear color pattern different from that of the rest of the mushroom images.
The correctly labeled mushrooms are all red, while the mislabeled ones are white or orange.
This finding indicates color is a critical feature that the CNN has learned to classify this class of images.

\subsection{Exploring Weight-Relevant Information}
\label{sec:case2}
\begin{figure}[htbp]
	\centering
	\includegraphics[width=0.98\textwidth]{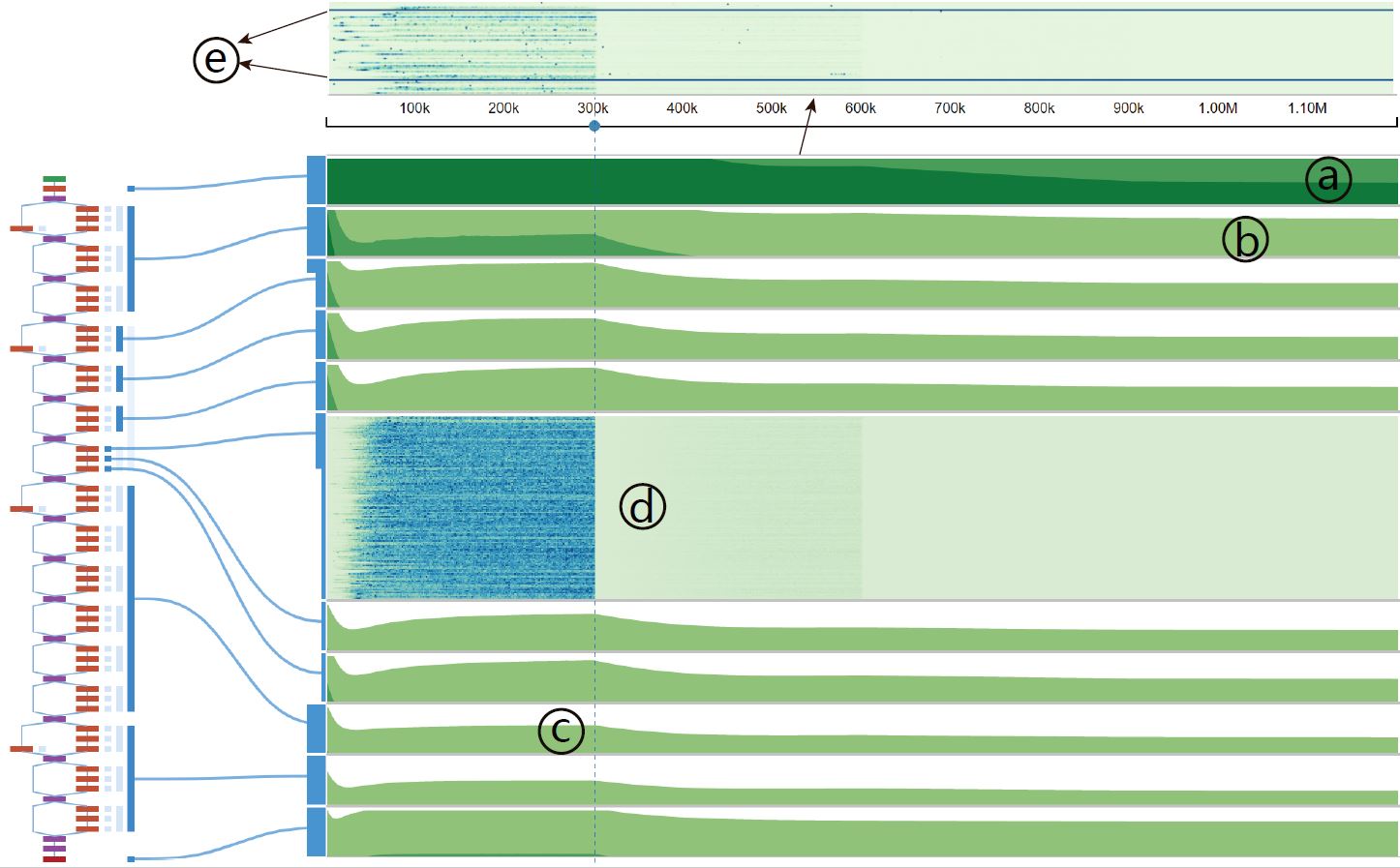}
	\caption{(a, b, c) The sd values of each layer decrease slowly and have different scales in different CONV modules. (d) The weight changes in filters are large at the beginning. (e) One outlier filter is detected, whose weights never change during the entire training process.}
	\label{fig:case2-1}
	\vspace{-2mm}
\end{figure}
This scenario shows how to discover patterns in neuron weights via the Layer View (\textbf{R1}, \textbf{R2}).




First, the experts choose to show the sd (standard deviation) of the weights at the layer-level using horizontal graphs (Fig.~\ref{fig:case2-1}).
As the experts expected, all the trends show a similar pattern of slow decrease, indicating the weights in the entire model is converging over iterations.
Besides, the experts also find that deeper layers (closer to the loss layer) tend to have smaller sd values.
\dy{In particular, by tuning the band number (finally to 3) of the horizon graphs, they found the sd values of a CONV module are usually twice as large as those of the one below it (a, b, and c in Fig.~\ref{fig:case2-1}).
Given that we apply Xavier initialization\footnoteref{note7} and for ResNet-50, the input sizes of layers in a CONV module are twice as large as the ones in the layers of its previous CONV module, the observed result is not beyond the experts' expectation. This suggests that there is no problem exist on the initialization approach.
}

\begin{figure}[t]
	\centering
	\includegraphics[width=0.98\textwidth]{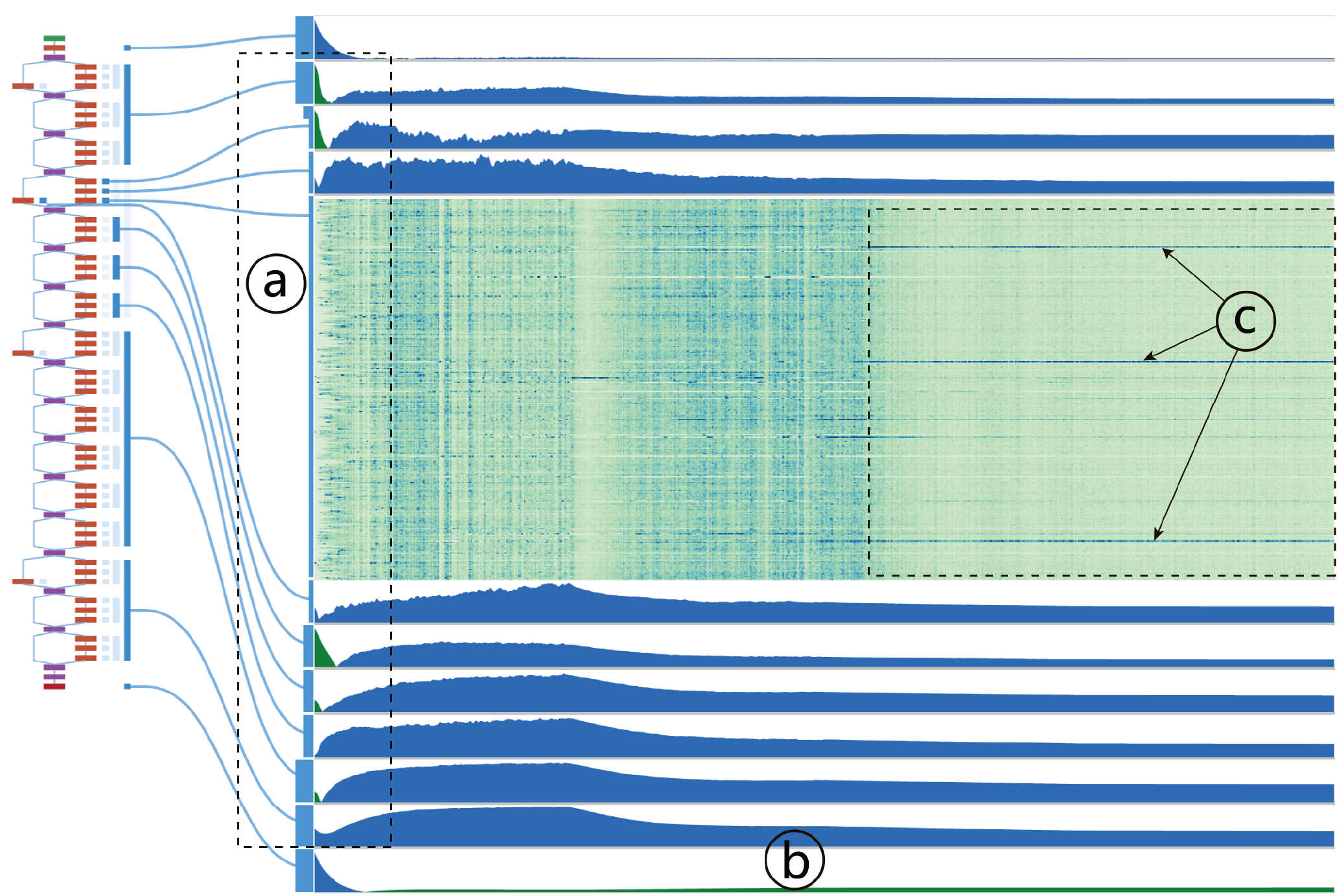}
	\caption{(a, b) The means of weights in each CONV layer become negative quickly (from green to blue) except for the FC layer. (c) Three filters are always more actively changed than the other filters in the later part of training progress.}
	\label{fig:case2-2}
	\vspace{-4mm}
\end{figure}

Analogously, the experts find that the weight means of each layer become negative quickly (from green to blue instantly, Fig.~\ref{fig:case2-2}a) except for the FC layer (Fig.~\ref{fig:case2-2}b).
At first, the pattern looks strange to the experts.
Then, the experts realize that it is reasonable to have more negative weights than positive ones, since negative values are often used to filter out trivial features owing to the use of ReLU activations.
The increase of negative weights suggests that the network is trained to extract useful information from the original image layer by layer, and then finally remain the most relevant features.
\dy{As for the FC layer, it plays a function of shaping the extracted useful features into feature vectors of given dimension for further classification. One strange phenomenon intrigues the experts, that is, the FC layer weight means are always positive in many-times training ResNet-50 (with different batch sizes and learning rates) on ImageNet Dataset, whereas becoming negative when training ResNet-164 on Cifar Dataset~\cite{krizhevsky2009learning}. This finding is worth a further investigation.}
Apart from layer-level values, the experts also explore the filter-level information (\textbf{R1}).
In our system, two different ways (i.e., filter-based or iteration-based) are used to normalize weight changes at the filter-level.
For filter-based normalization, changes are grouped and normalized by filters, which aims to help experts see the change distribution over iterations for individual filters.
Similarly, iteration-based normalization allows experts to examine the distribution over filters for individual iterations.
For example, Fig.~\ref{fig:case2-1}d visualizes the filter changes in one of the CONV layer belonging to the second CONV module using filter-based normalization.
The experts find that the changes are drastic in stage $s1$ and become relatively small in the later stages because of the decrease in learning rate and the convergence of the model.
However, the experts also identify two strange filters among 64 filters in the first CONV layer that have a constant deep blue color (Fig.~\ref{fig:case2-1}e).
By further checking, the experts find that the weights of these two filters never change during the entire training process.

\dy{This is a total surprise to the experts. Excluding programming bugs, the most likely reason should be due to the dying-ReLU problem, namely, these two filters are inactive for essentially all inputs and no gradients flow backward through the neurons of the two filters.  The experts suspect the dying-ReLU problem results from the high learning rate in the beginning of train. In fact, the experts usually follow a rule of thumb to set the hyper-parameter learning rate, that is, multiply the learning rate by k if the batch size is multiplied by k. This rule is currently formally introduced in a recent work~\cite{goyal2017accurate}.
In this experiment, we use 32 times larger batch size GPUs (32 GPUs) than the mini-batch size 32 for one GPU to train the model with the corresponding size of learning rate, dying-ReLU problem still occurs. This reminds the experts that the rule may not so accurate for extremely large batch size sometimes, but the problem can be solved by carrying out a warmup strategy (i.e., using lower learning rate at the start of training~\cite{he2016deep}), which the experts haven't done in previous trainings. One further interesting finding is that by inactivating these two "dead" filters (i.e., set their weights as 0 so that they are inactive for any inputs), the experts find the overall performance not affected at all, whereas if we inactivate other random-picked filters in the first CONV layer of the model, the number of mislabeled images in $D_v$ would increase few thousands. 
Thus, the experts finally modified the network configure and eliminated these two filters, so that the model can run faster while costing less memory.
}

Fig.~\ref{fig:case2-2}c visualizes the weight changes in one middle layer using iteration-based normalization.
The experts find that a small number of filters are always more actively changed than the other filters (long deep blue lines in Fig.~\ref{fig:case2-2}c) in the later part of iterations.
This pattern implies that the updates inside a layer may be highly divergent.
\dy{In the later part of the training, where the learning rate is getting smaller and the model is converging, only a couple of filters are still continually actively updated for every iteration. 
We have tried to inactivate these identified filters, the result showing that the overall performance is not affected.
It is not beyond the experts' expectation due to the ResNet's ensemble-like behavior~\cite{veit2016residual} (even several entire layers can be removed without impacting performance). In the end, the experts still cannot fully explain the behavior of these continually updating filters. One possible reason could be that these special filters are not trained well (not converge) to extract some specific features, thus reacting violently for every iteration even in the later stages of the training.}

\subsection{Exploring Filter-Image Correlations}
In this scenario, we demonstrate how the experts use the Correlation View to explore correlations between images and filters.

\begin{figure}[htb]
	\centering
	\includegraphics[width=1\columnwidth]{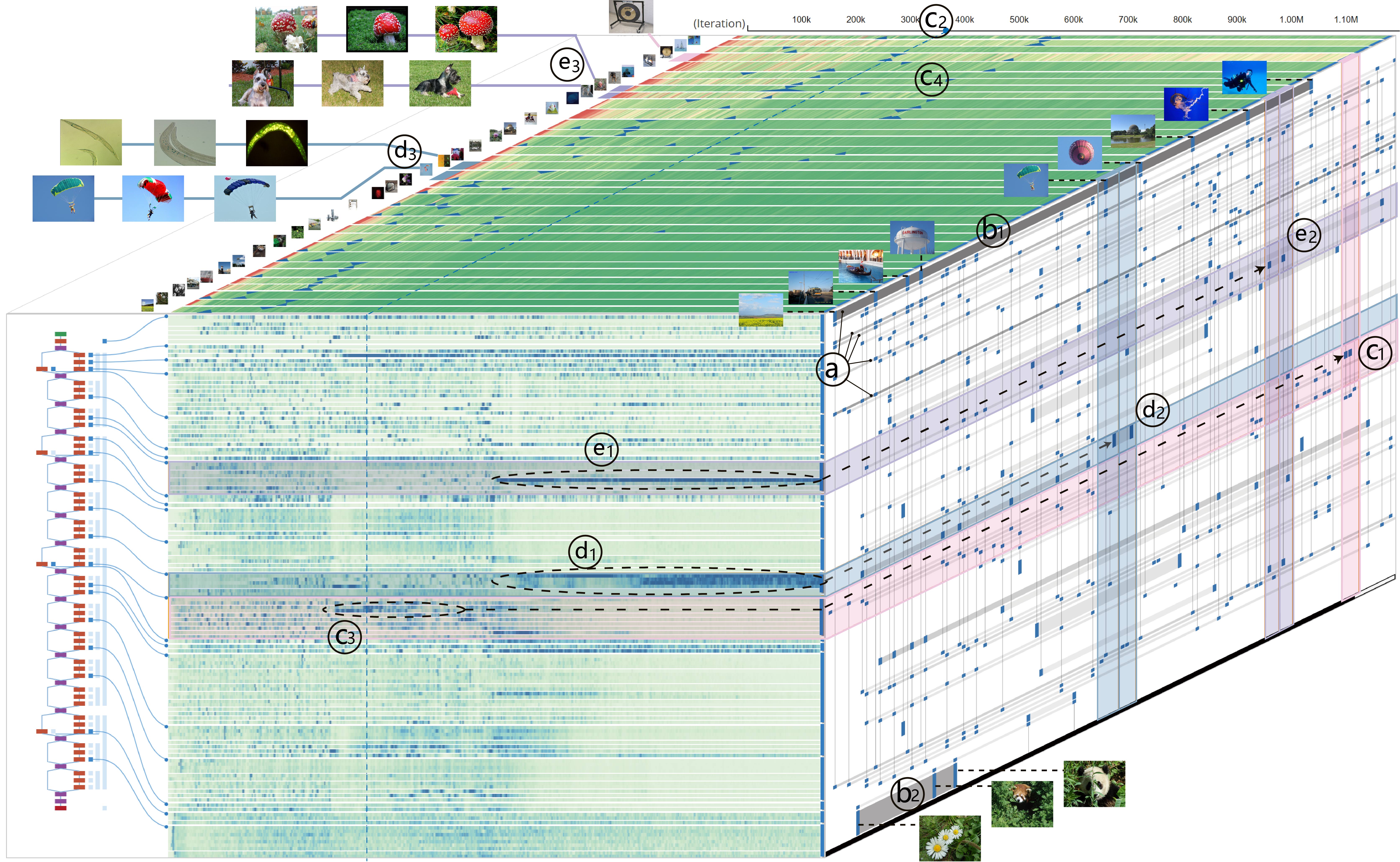}
	\vspace{-4mm}
	\caption{
		A cube-style visualization that fuses three coordinated views together to reveal the rich dynamics in a CNN training process:
		(top) the Validation View shows the error rate changes of validation classes; (front) the Layer View shows the weight changes in CNN
		filters; (right) the Correlation View shows the potential relationships between filters and validation classes.
	}
	\label{fig:teaser}
	\vspace{0mm}
\end{figure}

\textbf{Shallow-layer filters vs. deep-layer filters}.
\ti{At first, the experts choose to only show the top $k$ ($100$ in this case) changing filters in the layer view. By checking the network structure visualization, the experts find that the activated shallow layers (the layers close to data input layer) are more than the activated deep layers, and most activated layers are the last basic CONV layers of bottlenecks for deep CONV modules.
Besides, Fig.~\ref{fig:correlation_view}a shows that deep CONV modules tend to contain more anomaly filters (especially for the CONV modules 4). The experts think that this kind of knowledge is of great importance for network compression~\cite{han2015deep}.
Then, the experts go to the complicated version to examine the more detailed correlation information.
They filter the mini-sets with very few appearing times, finding that anomaly filters in shallow layers are generally shared by more anomaly classes (columns) and iterations (vertical lines in one column) than those in deep layers (Fig.~\ref{fig:teaser}a).
The experts think that this pattern may relate to the fact~\cite{veit2016residual} that shallow-layer filters are more likely to capture basic visual features than deep ones, thereby the huge change of these filters affecting more classes of images (e.g., the long and opaque lines marked by b in Fig.~\ref{fig:teaser}).
By contrast, a deep filter tends to learn higher-level features, thus only relating to specific classes of images.}

To further explore the correlations,
the experts select two mini-sets (b1 and b2 in Fig.~\ref{fig:teaser}), for comparison. Both the horizontal lines of b1 and b2 are opaque and thick.
By tracking them in the Layer View and the Validation View, the experts can see that b1 is in the first CONV layer, and related to many classes.
The experts open these classes and discover that many images in them have a common feature, i.e., a large background of blue sky or ocean (b1 in Fig.~\ref{fig:teaser}).
This discovery suggests that these filters in E1 may target the basic pattern to help identify images that contain large blue areas.
By contrast, b2 is located at the fifth \dy{CONV module} and related to only three classes.
Interestingly, the images in the three classes also share a more concrete feature, i.e., objects in a bush (b2 in Fig.~\ref{fig:teaser}).
\dy{In short, this case confirms that we are on the right track to reveal how the model weight changes relate to the classification result changes.}


\textbf{Important filters for a class}.
To find stronger correlations between filters and classes, the expert focus on anomaly filters that appear more than once in a cell for a specific class.
For example, the experts find two appearances of the same mini-set (containing two anomaly filters) for the class of ``gong'' (c1 in Fig.~\ref{fig:teaser}).
Tracking horizontally (along with the pink-highlighted area), the experts find that the mini-set does not appear in other anomaly iterations, which also implies a strong correlation between filters in the mini-set and the class.
Then, the experts click on these two rectangular glyphs to highlight the corresponding iterations on the timeline (c2 in Fig.~\ref{fig:teaser}) and the filter locations in the Layer View (c3 in Fig.~\ref{fig:teaser}).
It is clear that the gong class is not a well trained class as it has a very large yellow area (indicate a relatively high error rate) in the Validation View.
However, the experts also find a period in the middle when this class has a relatively good performance (c4 in Fig.~\ref{fig:teaser}), which happens to contain the highlighted anomaly iterations.
Meanwhile, the Layer View shows that the highlighted filters are also updated dramatically during the period of good performance (c3 in Fig.~\ref{fig:teaser}).
Considering these patterns, the experts speculate that the filters in the mini-set have a strong impact on the classification of gong images. 
As expected, we conduct experiments to inactivate these two filters, finding the overall performance and the performance on ``gong'' class are not impacted (see the reason in the last paragraph in Sec.~\ref{sec:case2}). Nevertheless, it provides the experts with a new manner to investigate the functions of filters co-working together for classifying one class of images. That is, increase the threshold to find anomaly filters as many as possible, find the mini-sets containing many filters to some classes from multiple layers, and then inactivate them all to validate corresponding impacts.

\textbf{Abnormal anomaly filters}.
The experts are also attracted by two mini-sets (d1 and e1 in Fig.~\ref{fig:teaser}), because of their abnormal color patterns.
The filters in these two mini-sets exhibit large changes all the time in the latter part of the training, which are very different from the other anomaly filters.
Thus, the experts are interested in these filters and further check their correlated classes in the Correlation View (right).
Interestingly, each abnormal mini-set only appears with two classes (d2 and e2 in Fig.~\ref{fig:teaser}), and each pair of classes have very similar performances displayed in the Validation View (d3 and e3 in Fig.~\ref{fig:teaser}).
By checking the detailed images of these classes, the experts discover some common patterns.
For example, for mini-set e1, the corresponding classes are about mushrooms growing on grass and dogs playing on grass (e3 in Fig.~\ref{fig:teaser}).
For mini-set d1, the corresponding classes are related to curved shapes, such as parachutes and round textures (d3 in Fig.~\ref{fig:teaser}).
Although the experts are still unclear about why these two mini-sets have such a special behavior, they believe that these filters are likely to play important roles in identifying middle-level features such as grass and curved shapes.
We also conduct further experiments to validate the impact of inactivating these filters and the results are similar to the previous case (i.e., important filters for a class).

\section{Expert Feedback}
\textbf{Usability} \dy{\name is built upon a close collaboration with three domain experts, who constantly underscore their requirements and provide suggestions during the implementation process.
After several iterations of refinement, the experts were happy with the current version. They all praised our way of effectively exploring such extreme large-scale training log via a hierarchical manner. 
\ea and \eb mentioned that the well-designed validation and layer views were very intuitive and helped them greatly. For example, the layer view allowing the experts to effectively observe and compare layer-related information (e.g., weight/gradient distribution) can help them diagnose network structures. The detecting of dying-ReLU problem in the early stage of a training is useful for tuning the hyper-parameters (e.g., learning rate). This kind of knowledge can also be leveraged to conduct model compression~\cite{han2015deep}, so as to improve the model in respect to computing speed and memory cost. Although the experts still cannot figure out the exact reason that some filters are always more actively updated in the later training stages, they believe the insight that would be obtained from the future investigation will be helpful in diagnosing and improving network structures. Besides, the divergent evolving patterns of classes and the numerous anomaly iterations found in validation view provide the experts with a new promising direction to train a better model.
Both \ea and \ec were particularly fond of the cube-style visualization and deemed it as a new perspective to observe the training of CNNs for them. 
They both have found many interesting patterns with the cube visualization, some of which were reasonable or could be explained after thinking for a while.
However, the experts also failed to figure out some other patterns, notwithstanding they conducted several testing experiments. 
Nevertheless, they were still excited that our system can help them identify potential subjects for further study.}



\textbf{Generality} During the implementation, we were concerned about the generality of \name, that is, whether the design was biased to the specific requirements from these three experts.
Therefore, to check how our system is accepted by broader expert communities, we presented our system in a workshop, which involved about 20 experts in the machine learning and visualization fields.
In addition, we also interviewed another group of twelve experts, who worked on a large project about using CNNs to improve image search quality.
We presented the latest version of \name to the experts, encouraged them to experiment with the system, and collected their feedback in the process.
Exceeding our expectation, \name was well accepted by these experts.
Although they proposed several new requirements, the experts shared many major interests with our three collaborators, such as tracking class level performance, filter-level information, and the relationships between them.
After introducing our system, they immediately understood the purposes of each view and all appreciated this novel, intuitive, and expressive way to watch training processes.
Although the demo was performed on our experiment datasets, the experts saw its potential in their project and immediately asked for collaboration with us, so that they could plug in and analyze their own datasets.

\textbf{Improvement} Apart from this positive feedback, the experts also made several interesting suggestions to further improve \name.
For example, two experts suggested that our current system only differentiates correct or incorrect classifications for validation images (i.e., 1 and 0).
However, the exact incorrect labels should also be presented because such information can help identify confusing or similar classes.
One expert mentioned that he showed strong interest on what happens before the anomaly iteration and suggested dump data of every iteration at that abnormal interval for fine-grained analysis.
Another expert suggested that our system shoud be integrated with online dashboards, as visualizing the training log on the fly can allow them to terminate the training early and save time if the visualization shows anything undesired.


\section{Discussion}
\label{sec:discussion}




\name is our first step to open the ``black box'' of CNN training.
Although the experts have high expectations of this tool, we all agree to start with two fundamental pieces of information: neuron weights and validation results.
Considering our target users and the large scale of datasets, we try to avoid using sophisticated visual encodings to ensure a fluent exploration experience.
Unsurprisingly, our bare-to-metal visualizations are preferred by the experts, and they use it to find many patterns easily, either expected or unexpected.
However, we still have several limitations.

First and foremost, although our system can effectively help experts identify strange or interesting patterns, there is still a gap between finding patterns and accelerating CNN training.
The experts still have to reason about and understand what these patterns mean or how to use them to accelerate model training in future.
We think it is not a problem faced just by our system, but by all CNN visualizations in general.
Like previous work, \name may only peel a hole in the box and reveal limited information.
But we hope that, by peeling enough holes, all these strange patterns will start to connect and make sense by themselves, thus providing a clear picture of CNNs.


\ti{Second, we have adopted many space-efficient visualizations and interaction techniques (e.g., hierarchy, filtering, and aggregation) to address the scalability issue.
Our current design can well support showing dozens of layers and classes in the same time. The correlation view shares all the filter strategies with the other two views, and vice versa.
Thus, our system can perform well in most cases.
Nevertheless, the worst scenario still requires to display hundreds or thousands of small multiples at the same time. 
A possible solution is to employ task-specific aggregation or filtering methods to show the data of interests.}

Third, we propose a rule-based anomaly detection method that requires experts to manually pick a reasonable window size $k$ and set the threshold for filtering. The number and patterns of anomalies are sensitive to these settings. One potential solution to this problem is to develop an automatic method to enumerate all potential parameter settings and identify those can detect a reasonable amount of significant anomalies and provide these settings to the experts as guidance.

Finally, we only conduct experiments on ResNet-50~\cite{he2016deep}, but our method can also be applied to other state-of-the-art deep CNN models, which often have similar hierarchical structures (e.g., ``inception block'' in google-inception-v4~\cite{szegedy2016inception}).
\ti{
Besides, the cube visualization is a general technique, which can be used to explore multiple heterogeneous time series data and their complex correlations.
However, to further generalize it, a strict user study has to be conducted to find the best manner to use it, such as the axis skew degree, and the minimum height/width for each row/column in the three faces.
}

\section{Conclusion}
\label{sec:conclusion}

We propose a novel visual analytics solution to disclose the rich dynamics of CNN training processes. Knowing such information can help machine learning experts better understand, debug, and optimize CNNs. We develop a comprehensive system that mainly comprises the validation, layer, and correlation views to facilitate an interactive exploration of the evolution of image classification results, network parameters, and the correlation between them.
We conduct experiments by training a very deep CNN (ResNet-50) on ImageNet, one of the largest labeled image datasets that is commonly used in practice, to demonstrate the applicability of our system. The positive feedback from our collaborating experts and the experts from an internal workshop validates the usefulness and effectiveness of our system.



Future studies may integrate some feature-oriented visualization techniques, which typically require recording the activation information for input instances. Feature visualizations can provide insights on what features a filter in a given snapshot of CNN has learned. Our system can track critical iterations to take snapshots for a CNN over training, and then use feature visualization techniques to analyze the learned features evolving patterns for the detected important filters. The other urgent need is to deploy the system in a real-time environment. To this end, we have to consider some new design and interaction requirements to fill the gap between finding patterns and accelerating CNN training.

%
%
%
%

\begin{acks}
	
	The authors would like to thank Kai Yan for providing support in editing the relevant media materials.
	
	The work is supported by the National Basic Research Program of China (973 program) under Grant No. 2014CB340304
	and ITC Grant with No. UIT/138.

\end{acks}

\bibliographystyle{ACM-Reference-Format}
\bibliography{reference}

\end{document}